\documentclass[letterpaper, 10 pt, conference]{ieeeconf}  

\IEEEoverridecommandlockouts                              

\usepackage{cite}
\usepackage{amsmath,amssymb,amsfonts}
\usepackage{bm}
\usepackage{algorithmic}
\usepackage{mathtools}
\usepackage{graphicx}
\usepackage{textcomp}
\usepackage{xcolor}
\usepackage{booktabs} 
\usepackage{scalerel,stackengine}
\usepackage[per-mode=symbol]{siunitx}
\stackMath
\newcommand\reallywidehat[1]{%
\savestack{\tmpbox}{\stretchto{%
  \scaleto{%
    \scalerel*[\widthof{\ensuremath{#1}}]{\kern-.6pt\bigwedge\kern-.6pt}%
    {\rule[-\textheight/2]{1ex}{\textheight}}
  }{\textheight}%
}{0.5ex}}%
\stackon[1pt]{#1}{\tmpbox}%
}

\usepackage{color}

\title{\LARGE \bf Online Trajectory Optimization for Dynamic Aerial Motions of a Quadruped Robot}

\author{Matthew Chignoli$^{1}$ and Sangbae Kim$^{1}$
\thanks{$^{1}$Department of Mechanical Engineering, Massachusetts Institute of Technology, Cambridge, MA 02139, USA: {\tt\small chignoli@mit.edu}} }%

\begin{document}

\maketitle
\thispagestyle{empty}
\pagestyle{empty}

\begin{abstract}

This work presents a two part framework for online planning and execution of dynamic aerial motions on a quadruped robot. Motions are planned via a centroidal momentum-based nonlinear optimization that is general enough to produce rich sets of novel dynamic motions based solely on the user-specified contact schedule and desired launch velocity of the robot. Since this nonlinear optimization is not tractable for real-time receding horizon control, motions are planned once via nonlinear optimization in preparation of an aerial motion and then tracked continuously using a variational-based optimal controller that offers robustness to the uncertainties that exist in the real hardware such as modeling error or disturbances. Motion planning typically takes between 0.05-0.15~\si{\second}, while the optimal controller finds stabilizing feedback inputs at 500~\si{\hertz}. Experimental results on the MIT Mini Cheetah demonstrate that the framework can reliably produce successful aerial motions such as jumps onto and off of platforms, spins, flips, barrel rolls, and running jumps over obstacles.

\end{abstract}

\newcommand{\Real}{\mathbb{R}}
\newcommand{\omegaBdHat}{\widehat{^B\bm{\omega}^d}}

\section{Introduction}
Replicating the dynamic capabilities of legged animals has long been the aim of researchers studying robot locomotion. Much of that research has focused on locomotion via standard walking and running gaits, specifically for bipedal and quadrupedal robots. This work has yielded impressive results, ranging from high speed running~\cite{di2018dynamic}, rough terrain locomotion~\cite{villarreal2020mpc, jenelten2020perceptive}, and vision-aided exploration~\cite{kolvenbach2020towards,brandao2020gaitmesh}. Less attention, however, has been devoted to using legs as a means for propelling robots into highly dynamic aerial behaviors, specifically behaviors involving significant rotational components (e.g. spins, barrel rolls, and flips). Dynamic aerial motions are a crucial step in expanding not only the set of environments and obstacles that robots are able to traverse, but also the tasks they are able to perform.

Research involving single leg hoppers is rather expansive, focusing on both control algorithms~\cite{yim2020precision,shen2020optimized} as well as leg and actuator designs~\cite{zhao2013msu,arikawa2002design}, but implementations of these findings on multi-leg robots capable of general locomotion behaviors is much less prevalent. In this work, we are instead interested in expanding the dynamic repertoire of medium to large scale quadruped robots capable of general locomotion tasks. Existing work involving aerial behaviors of larger scale legged robots has been conducted, but has typically been constrained to planar motions that, with the exception of~\cite{park2015online}, start from a relatively static initial position~\cite{xiong2018bipedal,wang2019untethered}. More complicated and dynamic aerial motions have been demonstrated on Boston Dynamics' Atlas robot, but these results have been limited to simulation~\cite{dai2014whole} or shown only in videos without details on control algorithms.

\begin{figure}[thb]
    \centering
    \includegraphics[width=180pt]{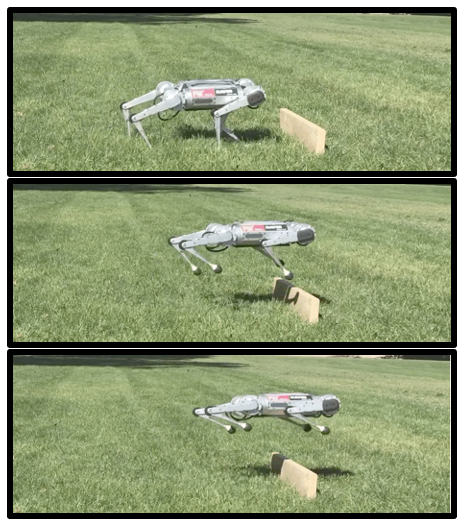}
    \caption{MIT Mini Cheetah executing a running jump from a trot gait over a 14 \si{\centi\meter} hurdle.}
    \label{fig:trot-jump}
\end{figure}

Dynamic aerial behaviors, such as the jump from a trotting gait shown in Fig.~\ref{fig:trot-jump}, are most commonly generated using trajectory optimization. Finding a suitable formulation that can handle challenges like non-flat terrain, significant body rotations, and contact constraints while maintaining tractable solve times is a complicated trade-off. While nonlinear optimal control using full robot dynamics has been demonstrated for quadruped locomotion~\cite{neunert2018whole}, many controllers instead use the more tractable centroidal dynamics of the robot when planning motions~\cite{winkler2018gait,ponton2016convex}. Centroidal dynamics can be optimized simultaneously with joint trajectories~\cite{dai2014whole} to ensure the dynamics of the optimization are equivalent to the full dynamics of the system. Despite the more tractable dynamics, the nonlinear optimizations required by these motions planners are still generally too slow to be solved in real time for most highly dynamic applications like jumping.

If continuous replanning cannot be done online, trajectories must be generated either offline or once online and then tracked using a feedback controller. Linear quadratic regulator (LQR)-based controllers are common for this tracking task, yet many existing controllers used to stabilize trajectories for high degree of freedom robots either use hightly simplified dynamics like zero moment point~\cite{kuindersma2014efficiently}, or do not consider the robot's body orientation in the linearized state~\cite{herzog2015trajectory}. Neither of these simplifications is acceptable for dynamic aerial motions that might include large rotations of the body. Including orientation dynamics in the linearized state of a reduced order quadruped model has been achieved using variational-based linearization~\cite{wu2015variation,chignoli2020variational,ding2019real}, a technique that parameterizes error on the manifold $SO(3)$ for convenient use in a quadratic cost function while also avoiding issues like singularities and unwinding, which plague Euler angles and quaternions, respectively.

The primary contribution of this work is a framework for reliably generating and tracking dynamic aerial motions for a quadruped robot, including motions with large rotational components. The framework combines centroidal momentum-based trajectory optimization with variational-based optimal control to reliably perform leaps, spins, flips, and running jumps generated in real-time. The trajectory optimization is able to handle the trade-off between accurate dynamics and real-time solvability, while the optimal controller is able to stabilize trajectories in a way that is robust to inevitable discrepancies such as modeling errors in the optimization model or perturbations to the robot before or during takeoff. Specifically, we demonstrate that the framework can be used on the MIT Mini Cheetah to generate jumps online onto tables 1.3 times the standing height of the robot and over gaps wider than the length of the robot's body.

The rest of this paper is organized as follows. Section II outlines the overall approach for planning and tracking aerial motions. Section III gives implementation details including specifics about generating reference trajectories and solving the relevant optimal control problem. Section IV presents the results of successful implementation of the framework on the MIT Mini Cheetah, and Section V discusses conclusions from the work.

\section{Online Jumping Framework}
\begin{figure*}[t]
    \centering
    \includegraphics[width=500pt]{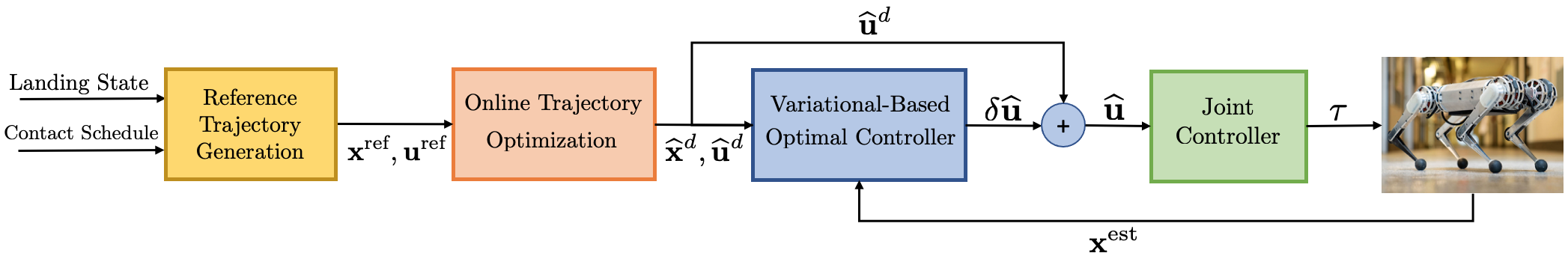}
    \caption{Framework for online generation and tracking of dynamic aerial motions. Online trajectory optimization runs once in preparation of an aerial motion, followed by the tracking and joint controllers running at 500~\si{\hertz}.}
    \label{fig:block-diag}
\end{figure*}

In order for legged robots to be deployed reliably in challenging, real-world environments, they will need to be able to generate novel motions, like jumping, online that allow them to conquer obstacles of arbitrary size, orientation, and location relative to the robot. The generality of a framework that produces such a rich set of motions for novel scenarios is improved as the amount of task specific cost function tuning, warm starting, and assumptions about the model and world is reduced. The approach outlined in this section and illustrated in Fig.~\ref{fig:block-diag} provides such a framework for not only producing novel motions online, but also robustly tracking these motions in a way that enables reliable performance on the MIT Mini Cheetah.

\subsection{Trajectory Optimization}
The online trajectory optimization stage of the framework uses a centroidal dynamics-based model of the robot 
to plan aerial motions. The model is simplified according to common assumption that the robot can be treated as a single rigid-body (SRB). This means that the legs and torso of the robot are lumped together as a single body with fixed inertia. Consequently, the joint level dynamics and kinematics of the robot are not included in the optimization. Rather it is assumed that as long as the feet remain within an approximate Cartesian workspace relative to their respective hips, they will not violate kinematics limits or collide with one other. The state of the SRB model of the robot is given by 
\begin{equation}
    \mathbf{x} = \begin{bmatrix} \mathbf{p}_c^T & \mathbf{\Theta}^T & \mathbf{v}_c^T & ^\mathcal{B}\bm{\omega}^T \end{bmatrix}^T, \label{SRBMstate}
\end{equation}
where $\mathbf{p}_c\in\Real^3$ is the position of the center of mass (COM), $\mathbf{\Theta}\in\Real^3$ is the orientation of the SRB given by Euler angles, $\mathbf{v}_c\in\Real^3$ is the velocity of the center of mass, and $^\mathcal{B}\bm{\omega}\in\Real^3$ is the angular velocity of the SRB represented in the body-fixed frame $\mathcal{B}$. The SRB model state is controlled via the net wrench exerted on the robot's COM via the reaction forces at the feet. The net wrench is given by
\begin{equation}
    \begin{bmatrix} \mathbf{f}  \\ \bm{\tau} \end{bmatrix} = \sum_{i=0}^{n_c} \begin{bmatrix} \mathbf{f}_i  \\ (\mathbf{p}_i - \mathbf{p}_c) \times \mathbf{f}_i \end{bmatrix},
\end{equation}
where $\mathbf{f}\in\Real^3$ is the net force and $\bm{\tau}\in\Real^3$ is the net moment on the COM, $\mathbf{f}_i\in\Real^3$ the force at foot $i$, $\mathbf{p}_i\in\Real^3$ the position of foot $i$, and $n_c$ the number of contact points. Thus, the control input to the system can be represented
\begin{equation}
    \mathbf{u} = \begin{bmatrix} \mathbf{p}_1^T & \mathbf{f}_1^T & \hdots & \mathbf{p}_{n_c}^T & \mathbf{f}_{n_c}^T\end{bmatrix}^T.
\end{equation}

The state and control trajectories of the robot are discretized into $n_t$ timesteps with time $\Delta t$ between each step. The state and control trajectory variables in the optimization are given by $\mathbf{X}\in\Real^{12\times n_t}$ where $\mathbf{X} = [\mathbf{x}_1,\hdots,\mathbf{x}_{n_t}]$ and $\mathbf{U}\in\Real^{6n_c\times (n_t-1)}$ where $\mathbf{U} = [\mathbf{u}_1,\hdots,\mathbf{u}_{n_t-1}]$. 

The optimization finds the set of control inputs $\mathbf{U}$ that solves the following,
\begin{align}
& \underset{\mathbf{X},\mathbf{U}}{\text{minimize}}
& & \sum_{k=1}^{n_t-1}{\big(\mathbf{\Tilde{x}}_k^T\mathbf{Q}_k\mathbf{\Tilde{x}}_k + \mathbf{\Tilde{u}}_k^T\mathbf{R}_k\mathbf{\Tilde{u}}}_k\big)+\mathbf{\Tilde{x}}_N^T\mathbf{Q}_N\mathbf{\Tilde{x}}_N, \label{cost-fun}
\end{align}
where $\mathbf{\Tilde{x}}$ and $\mathbf{\Tilde{u}}$ are deviations from reference trajectories $\mathbf{x}^\text{ref}$ and $\mathbf{u}^\text{ref}$ and $\mathbf{Q}$ and $\mathbf{R}$ are weight matrices. The dynamics of the robot are enforced as constraints on the optimization via Euler integration of the robot's state between time steps
\begin{equation}
    \mathbf{x}_{k+1} = \mathbf{x}_{k} + \mathbf{\dot{x}}_k\Delta t, \label{dyn_con}
\end{equation}
where $\mathbf{\dot{x}}$ encodes the SRB model dynamics. These dynamics are given by
  \begin{equation}
     \dot{\mathbf{x}} = \frac{d}{dt}\begin{bmatrix}
    \mathbf{p}_c \\[.5ex]
    \mathbf{\Theta} \\[.5ex]
    \mathbf{v}_c \\[.5ex]
    ^\mathcal{B}\bm{\omega}
  \end{bmatrix} = \begin{bmatrix}
    \mathbf{v}_c \\[.5ex]
    \mathbf{B}(\mathbf{\Theta}) {^\mathcal{B}\bm{\omega}} \\[.5ex]
    \frac{1}{m}\mathbf{f} - \mathbf{g} \\[.5ex]
    ^\mathcal{B}\mathbf{I}^{-1}(\mathbf{R}_b^T\,\bm{\tau}-\,{}^\mathcal{B}\bm{\omega}\times \, {^\mathcal{B}\mathbf{I}} \, ^\mathcal{B}\bm{\omega)}
  \end{bmatrix}, \label{dynamics}
 \end{equation}
 where $\mathbf{B}(\mathbf{\Theta})\in\Real^{3\times 3}$ is an orientation-dependent matrix that converts angular velocity to Euler angle rates, $m$ is the total mass of the robot, $\mathbf{g}\in\Real^{3}$ is the acceleration of the body due to gravity, $^\mathcal{B}\mathbf{I}\in\Real^{3\times 3}$ is the inertia tensor about the CoM in the body-fixed frame, and $\mathbf{R}_b\in SO(3)$ is the rotation matrix from the body frame to the inertial frame. Since the legs and body are treated as a single rigid-body of constant mass and size, the rotational inertia of the body $^\mathcal{B}\mathbf{I}$ is assumed constant.
 
A further constraint imposed on the optimization is that when the $i$th foot is scheduled to be in contact with the ground at a given time step, the foot must be touching the ground,
\begin{equation}
    z_g(p_{x,i},p_{y,i}) - p_{z,i} = 0
\end{equation}
where $z_g$ gives the height of the ground at a given location. Each foot is constrained to remain in place throughout each planned stance phase. Additionally, the position of each foot relative to its respective hip is constrained by
\begin{equation}
    \vert\vert \mathbf{p}_{i} - \mathbf{p}_{hip,i} \vert\vert - \ell_{\text{max},i} \le 0,
\end{equation}
where $\ell_{\text{max},i}$ is the maximum length of the $i$th limb. Lastly, the ground reaction forces for each foot in contact are constrained to remain inside the friction pyramid $\mathcal{F}$, given by
\begin{equation}
    \mathcal{F}_i \coloneqq \{\mathbf{f}\in\Real^3 : \vert f_x\vert \le \mu_i f_z, \vert f_x\vert \le \mu_i f_z, 0 \le f_z \le f_{\text{max},i} \},
\end{equation}
where $\mu$ is the coefficient of friction between the ground and foot and $f_{\text{max}}$ is a limit on the allowable normal force. When a foot is scheduled to be out of contact, i.e. in swing, $f_{\text{max}}$ is set to zero.


The trajectory optimization returns an optimal control sequence of stance foot positions and ground reaction forces over the course of the takeoff period along with the corresponding state trajectory for the robot. This returned trajectory, however, is not guaranteed to be optimal, or even near optimal, for any initial state of the robot other than the one that was provided to the optimization. Similarly, there is no guarantee that if perturbed the trajectory will be stable or result in a velocity at liftoff near that of the optimal solution. Since the optimization is not fast enough to be continuously re-plan, we track the optimal trajectory using the variational-based optimal controller (VBOC) explained in Section II-B.

\subsection{Variational-Based Optimal Control}
Linear quadratic regulators are a popular option for tracking nonlinear reference trajectories, but two crucial aspects of legged locomotion prevent the application traditional LQR-based feedback control for tracking dynamic aerial motions. These challenges include contact constraints and the nonlinear orientation dynamics of floating base systems like legged robots. LQR in its typical form is ill-suited to handle constraints on control inputs, which in this case come from friction and positive normal force requirements. Furthermore, conventional linearization techniques are unable to linearize the orientation dynamics of floating base systems in 3D because they evolve on the non-Euclidean manifold $SO(3)$. 

In previous work~\cite{chignoli2020variational}, we proposed a controller that offers solutions to these two challenges. Variational-based linearization (VBL) exploits specific variation expressions on the manifold $SO(3)$ and can thus be applied to the time-varying optimal trajectories computed via the formulation in Section II-A. The result is a compact linear model for the system dynamics that is free from shortcomings like singularities or unwinding phenomena. Control constraints are handled via a two step cost-to-go approximation. The first step involves computing the cost-to-go of the given optimal control problem, assuming there are no constraints on the control input. Next, a quadratic program descends this ``unconstrained cost-to-go" from step one, while enforcing linear constraints on the control input to give an approximation of the constrained optimal policy. 

The VBL produces linear dynamics in the form
\begin{equation}
    \dot{\mathbf{s}} = \mathbf{A}(\mathbf{\widehat{x}}^d(t),\mathbf{\widehat{u}}^d(t))\mathbf{s}+\mathbf{B}(\mathbf{\widehat{x}}^d(t),\mathbf{\widehat{u}}^d(t))\delta \bm{u}, \label{linearSys}
\end{equation}
where $\mathbf{s}$ approximates the error between the actual state of the system, $\mathbf{\widehat{x}}$, and the reference state, $\mathbf{\widehat{x}}^d$, and $\mathbf{\widehat{u}}$ is a variation to the reference control input $\mathbf{\widehat{u}}^d$. It is important to note that the robot state and controls used in the VBOC, $\mathbf{\widehat{x}}$ and $\mathbf{\widehat{u}}$, need not be the same as the $\mathbf{x}$ and $\mathbf{u}$ used by the optimization in Section II-A. In this work, the VBOC state includes the positions of the robot's feet
\begin{equation}
    \mathbf{\widehat{x}} = \begin{bmatrix} \mathbf{x}^T & \mathbf{p}_1^T  & \hdots & \mathbf{p}_{n_c}^T  \end{bmatrix}^T,
\end{equation}
where $\mathbf{x}\in\Real^{12}$ is the SRB model state from~\eqref{SRBMstate} and $\mathbf{p}_i\in\Real^3$ is the position of the $i$th foot. The VBOC control input therefore includes only the reaction forces at the feet
\begin{equation}
    \mathbf{\widehat{u}} = \begin{bmatrix} \mathbf{f}_1^T  & \hdots & \mathbf{f}_{n_c}^T  \end{bmatrix}^T.
\end{equation}
The reason for this rearrangement is because our robot's swing leg tracking performance is rather poor. As a result, the actual placement of the robot's feet can deviate from the desired position by as much as 4.5~\si{\centi\meter} when trotting at 1~\si{\meter\per\second}. Including the foot position in the state allows the controller to reason about the effects that the foot position error has on the robot's dynamics.

The resulting quadratic program for the VBOC is given by~\cite{chignoli2020variational}
\begin{equation}
    \min_{\delta\mathbf{\widehat{u}} \in \mathcal{U}(\mathbf{\widehat{u}}^d)} \delta \mathbf{\widehat{u}}^T\mathbf{H}\delta \mathbf{\widehat{u}}+ 2\mathbf{s}^T\mathbf{P}\mathbf{B}\delta\mathbf{\widehat{u}}, \label{VBL-QP2}
\end{equation}
where $\mathcal{U}(\mathbf{\widehat{u}}^d)$ is a set of control input constraints dependent on the reference input, $\mathbf{H}$ is a weighting matrix, and $\mathbf{P}$ is the Riccati matrix in the unconstrained cost-to-go, $V_{unc}^* = \mathbf{s}^T\mathbf{P}\mathbf{s}$. The Riccati matrix can be computed via the algebraic Riccati equation for infinite horizon problems or via the Riccati differential equation for finite horizon cases~\cite{bertsekas1995dynamic}.

\section{Implementation Details}
To demonstrate the viability of the jumping framework outlined in Section II, we deploy it on the MIT Mini Cheetah quadruped robot~\cite{katz2019mini}. This section reviews implementation details relevant to the successful demonstration of the framework. Specifically, it outlines how reference trajectories were generated to create various aerial motions, as well as how the finite-horizon optimal control problem involved in the VBOC is handled.

\begin{table*}[thb]
\centering
\caption{Average Solve Times for Various Jumping Motions}
\label{tab:my-table}
\begin{tabular}{|c|c|c|c|c|c|c|}
\hline
Aerial Motion & $180^\circ$ Spin & \begin{tabular}[c]{@{}c@{}}34 cm Platform\\ Jump (Forward)\end{tabular} & \begin{tabular}[c]{@{}c@{}}15 cm Jump\\ Off w/ Spin\end{tabular} & \begin{tabular}[c]{@{}c@{}}20 cm Platform\\ Jump (Lateral)\end{tabular} & Trot Jump & Bound Jump \\ \hline
Horizon Length, $n_t$ & 13 & 13 & 13 & 13 & 17 & 17 \\ \hline
Time Discretization, $\Delta t$ (s) & 0.02 & 0.02 & 0.02 & 0.02 & 0.025 & 0.025 \\ \hline
Avg. Solve Time (ms) & 57.2 & 62.1 & 77.8 & 65.2 & 110.0 & 118.6 \\ \hline
Avg. Iterations to Converge & 10 & 11 & 12 & 11 & 18 & 18 \\ \hline
\end{tabular}
\end{table*}

\subsection{Reference Trajectory Generation}
All aerial motions in this paper emerge based on the combination of contact sequence and reference trajectory, $\mathbf{x}^\text{ref}$ and $\mathbf{u}^\text{ref}$, provided to the optimization. Through experimentation, the most simple yet effective method we found for selecting a contact sequence was to continue the pre-jump gait of the robot. For example, if you wish to produce a jump from a bounding gait, you can simply optimize over a single gait cycle by splitting the gait into $n_t$ discrete time steps. In this case, that single gait cycle constitutes the ''takeoff phase" of the jump. We found this simple method for contact sequence selection to be capable of generating 3D running jumps at speeds up to 1.3~\si{\metre\per\second} for gaits including pronking, bounding, and trotting. In the case of jumps starting from a static position, all feet are assumed to remain in their given position until the end of the takeoff phase, which typically lasts between 0.2-0.25~\si{\second}.

Simple cost functions can similarly produce rich sets of behaviors. A quadratic cost is used in the optimization that penalizes deviations from a reference trajectory. The trajectory optimization plans only the takeoff phase of each jump. Since the robot's momentum throughout flight is dictated by its momentum at the instant of liftoff, we formulate the objective function such that the terminal cost, $\mathbf{\Tilde{x}}_N^T\mathbf{Q}_N\mathbf{\Tilde{x}}_N$, dominates. Depending on the desired final position of the robot (e.g. jumping onto a table, over a hurdle, or flipped 360$^\circ$ about a desired axis), we use simple equations for projectile motion to determine the state at liftoff that produces the desired CoM trajectory and angular momentum throughout flight.

The desired position of the robot when it lands, $\mathbf{p}_c^\text{ref}(t_{LA})\in\Real^3$, is related to the reference trajectory of the robot throughout takeoff via 
\begin{equation}
    \mathbf{p}_c^\text{ref}(t_{LA}) = \mathbf{p}_{c}^\text{ref}(t_{LO}) + \mathbf{v}_{c}^\text{ref}(t_{LO})\Delta t_{FL} + \frac{1}{2}\mathbf{a}_g\Delta t^2_{FL}, \label{projectile}
\end{equation}
where $\mathbf{p}_{c}^\text{ref}(t_{LO})\in\Real^3$ and $\mathbf{v}_{c}^\text{ref}(t_{LO})\in\Real^3$ are the position and velocity of the robot when it lifts off the ground, $\Delta t_{FL}$ is the duration of the flight phase, and $\mathbf{a}_g\in\Real^3$ is the acceleration of the robot due to gravity.

Since the initial state of the robot is known, the reference position and velocity can be computed by integrating the reference acceleration $\mathbf{a}_{c}^\text{ref}(t)\in\Real^3$ of the robot from zero to the time of liftoff, $t_{LO}$. In this work, we generate reference trajectories by first manually authoring simple trajectories for the vertical component of the reference acceleration. Empirically, we found the following heuristic for vertical acceleration to work well
\begin{equation}
    a_{c_z}^\text{ref}(t) = (\beta + \frac{t}{t_{LO}}\gamma)\frac{1}{m}\sum_{i=1}^{n_c}\phi_{i}(t) f_{\text{max},i}, \label{aref}
\end{equation}
where $\beta\in\Real$ and $\gamma\in\Real$ are scaling parameters and $\phi_{i}(t)\in[0,1]$ indicates whether the $i$th foot is in or out of contact at time $t$. Using the trajectory for $a_{c_z}^\text{ref}(t)$, the reference trajectories for $v_{c_z}^\text{ref}(t)$ and $p_{c_z}^\text{ref}(t)$ can be obtained via integration from the initial state. Consequently, the vertical component of~\eqref{projectile} can be solved to obtain the duration of flight, $\Delta t_{FL}$, assuming a landing height of $p_{c_z}(t_{LA})$. Assuming constant acceleration in the forward and lateral directions, the forward and lateral components of~\eqref{projectile} can be solved to obtain the $a_{c_x}^\text{ref}$ and $a_{c_y}^\text{ref}$ that result in the robot landing at $\mathbf{p}_c^\text{ref}(t_{LA})$.

A similar process can be used to generate rotational motions like barrel rolls. Assuming the robot only rotates about a single principal axis, the rotation about that axis is given by
\begin{equation}
        \theta(t_{LA}) = \theta^\text{ref}(t_{LO}) + \omega^\text{ref}(t_{LO})\Delta t_{FL},  \label{proj-angle}
\end{equation}
where $\theta\in\Real$ is the angular displacement about the axis of interest, $\omega\in\Real$ is the angular velocity about this axis, and $\alpha\in\Real$ is the angular acceleration. Again, since $\Delta t_{FL}$ is known, the required constant acceleration $\alpha^\text{ref}$ can easily be computed via~\eqref{proj-angle} and used to generate the angular component of $\mathbf{x}^\text{ref}$.

The reference trajectory $\mathbf{x}^\text{ref}$ used in~\eqref{cost-fun} is built by integrating $\mathbf{a}^\text{ref}$ forward in time, and the reaction forces in $\mathbf{u}^\text{ref}$ are selected such that $i$th foot has reference force
\begin{equation}
    \mathbf{f}_{i}^\text{ref}(t) = \frac{\phi_{i}(t)}{\Phi(t)}\mathbf{a}_{c}^\text{ref}(t)
\end{equation}
where $\Phi(t)\in\Real$ is the total number of feet in contact at time $t$. Lastly, the foot placements in $\mathbf{u}^\text{ref}$ are selected such that the feet are under their respective hips at all times.

\subsection{Finite-Horizon Optimal Control}

While the balancing task demonstrated previously using VBOC~\cite{chignoli2020variational} was amenable to formulation as an infinite-horizon optimal control problem, the aperiodic nature of the aerial motions in this work necessitates their treatment as finite horizon problems. Consequently, the Riccati matrix in~\eqref{VBL-QP2} must be obtained via backward integration of the Riccati differential equation
\begin{equation}
\begin{split}
     -\dot{\mathbf{P}}(t) = \mathbf{Q}-\mathbf{P}(t)\mathbf{B}(t)\mathbf{R}&^{-1}\mathbf{B}(t)^T\mathbf{P}(t) \\ 
     &+\mathbf{P}(t)\mathbf{A}(t)+\mathbf{A}(t)^T\mathbf{P}(t) \label{CARE},
\end{split}
\end{equation}
with boundary condition $\mathbf{P}(t_f) = \mathbf{P}_f\in\Real^{24\times 24}$. This backward integration need only be carried out a single time after the initial trajectory optimization. The size of the quadratic program~\eqref{VBL-QP2} is therefore invariant to the horizon length of the reference trajectory, $\mathbf{\widehat{x}}_d$. This invariance is a feature not present in most model predictive-based controllers, and it allows the controller to run at a rate of 500~\si{\hertz} for trajectories of arbitrary duration.

The VBOC runs at 500Hz, which means that the nominal trajectory it linearizes about, $\mathbf{\widehat{x}}_d(t)$ and $\mathbf{\widehat{u}}_d(t)$, and the time-varying Riccati matrix $P(t)$ must be uniformly sampled every 0.002~\si{\second}. The trajectory optimization~\eqref{cost-fun}, however, typically uses a time discretization between 0.01-0.04~\si{\second}. To account for this discrepancy when sampling from the optimal trajectories, we linearly interpolate between the collocation points of $\mathbf{\widehat{x}}^d(t)$ and $\mathbf{\widehat{u}}^d(t)$. While, in general, resampling from low to high frequencies can induce instability, we found that linear interpolation of the optimal trajectory combined with a fourth-order Runge-Kutta scheme for integrating the Riccati equation~\eqref{CARE} produced stable Riccati solutions as well as robust tracking behaviors of the robot. 

\section{Results}
The framework is implemented successfully on the MIT Mini Cheetah both in simulation and on hardware. The nonlinear optimization described in Section II-A is solved using IPOPT~\cite{wachter2006implementation}, a the freely available NLP solver. The VBOC quadratic program~\eqref{VBL-QP2} is solved using the quadratic programming library qpOASES. Due to the limited onboard computing power of the MIT Mini Cheetah, nonlinear trajectory optimization using IPOPT is run off board on a laptop with an Intel i7 processor. All quadratic programs, however, are solved onboard the Mini Cheetah on its Intel UP Board computer.

The average solve times for the various aerial motions, shown in Table~\ref{tab:my-table} range from 50~\si{\milli\second} to 120~\si{\milli\second}. The same set of gains and cost function weights is used for all motions. The maximum normal force $f_\text{max}$ for all legs is set to 140~\si{\newton} and the coefficient of friction $\mu$ between the feet and ground is set to 1.5. For all jumps performed in this paper, landing is achieved via simple PD control of the joints to a fixed landing position. For the running jumps, these solve times dictate the minimum distance at which we must begin preparing for an obstacle.  For nominal running speeds of the Mini Cheetah (1-2~\si{\meter\per\second}), average solves times under 0.12~\si{\second} means the optimization can be triggered when a detected obstacle is as close as 12-24~\si{\centi\metre} ahead. 

\begin{figure}[thb]
    \centering
    \includegraphics[width=\columnwidth]{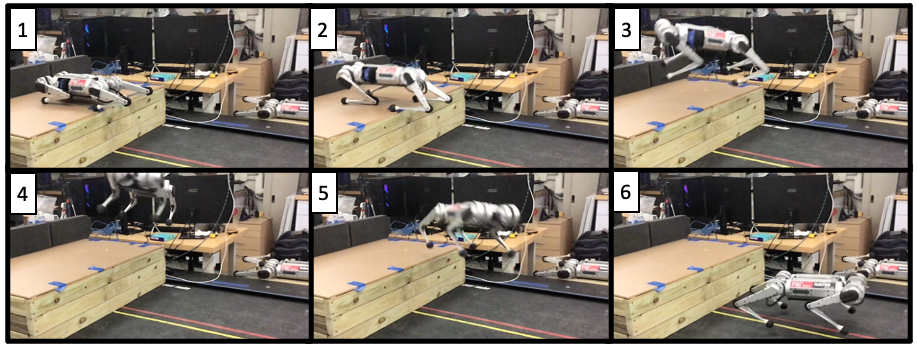}
    \caption{Hardware demonstration of the MIT Mini Cheetah performing a spinning jump off of a 26cm tall platform.}
    \label{fig:table-spin-hardware}
\end{figure}

\subsection{Static Jumps}
\begin{figure}[thb]
    \centering
    \includegraphics[width=\columnwidth]{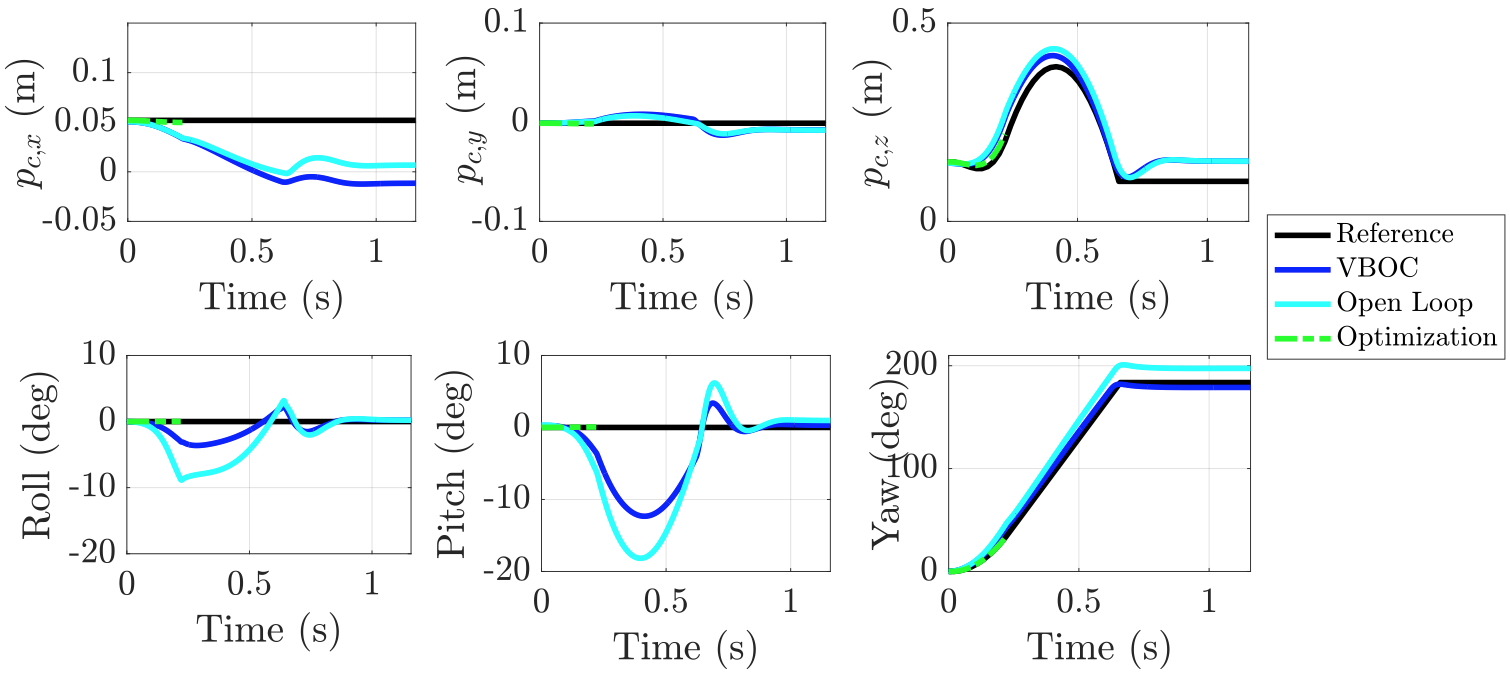}
    \caption{Comparison of the reference trajectory to the trajectory realized on the robot when performing a 180$^\circ$ spin about the robot's yaw axis.}
    \label{fig:mc-spin-VBOC}
\end{figure}

The framework was used to produce a variety of aerial motions with the robot starting from a static initial pose. These motions, which are all demonstrated in the accompanying video\footnote{https://www.youtube.com/watch?v=3cFSbVxAzCA}, include jumps onto and off of a table, 180\si{\degree} spins, barrel rolls, and combinations of these motions, such as the jump shown in Fig.~\ref{fig:table-spin-hardware}. Simulation results comparing the performance of open loop ground reaction force commands versus trajectory tracking using variational-based optimal control are shown in Fig.~\ref{fig:mc-spin-VBOC}, in which the robot performs a 180$^\circ$ spin about its yaw axis. The commanded reaction forces for the open loop case are obtained via linear interpolation of the optimized forces $\mathbf{\widehat{u}}^d$.

The results in Fig.~\ref{fig:mc-spin-VBOC} prove not only that the jumping controller can effectively perform the desired aerial rotation, but also that the VBOC adds considerable stability to the jump. The superior performance of the VBOC is a result of its ability to account for model discrepancies of the SRB model as well as its consideration of the effect that current actions will have on future cost. This consideration of future costs is encoded via the Riccati matrix $P$ in ~\eqref{VBL-QP2}.

\begin{figure}[thb]
    \centering
    \includegraphics[width=\columnwidth]{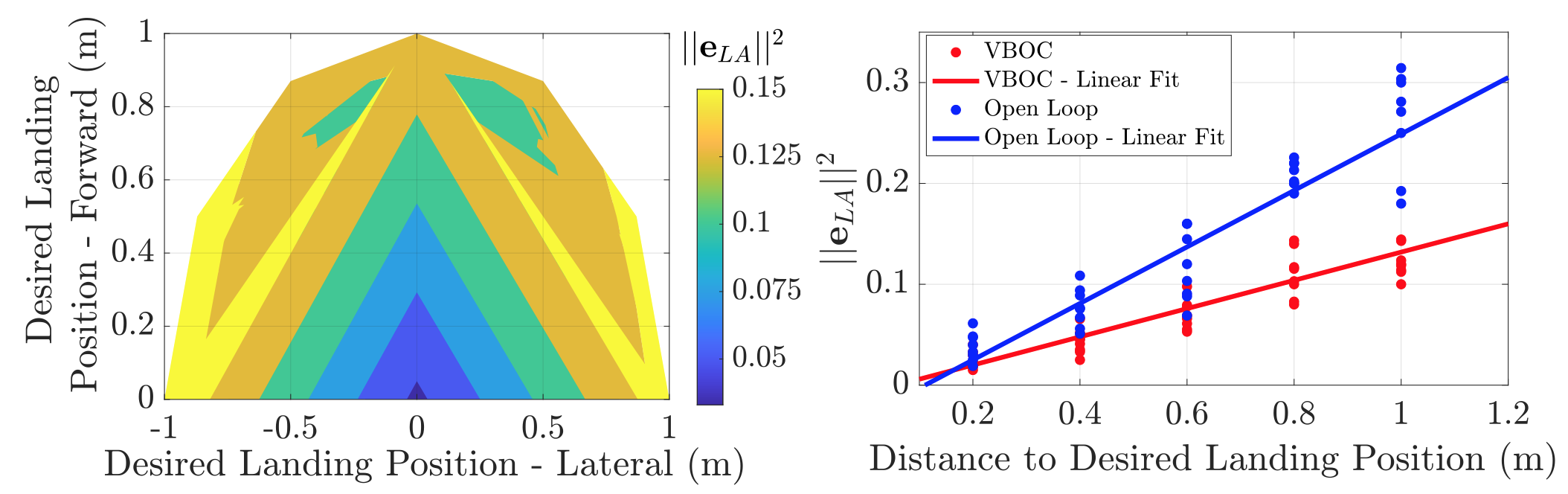}
    \caption{Analysis of the average error between the desired landing location of the robot $\mathbf{p}_c^\text{ref}(t_{LA})$ and the acheived landing location of the robot $\mathbf{p}_c(t_{LA})$.}
    \label{fig:landing}
\end{figure}

Despite the simplicity of our approach used to generate reference trajectories for the optimization, we demonstrate that the optimization is able to closely track these reference trajectories and produce jumps that land reliably close to the desired landing location. The results of 40 simulated jumps with various desired landing positions are shown in Fig.~\ref{fig:landing}. For each jump, $\mathbf{a}_{c}^\text{ref}(t)$ is generated from~\eqref{aref} using $\beta=-0.1$ and $\gamma=0.45$. The error metric used in these plots, $\mathbf{e}_{LA}\in\Real^3$, is calculated based on the difference between the desired landing position $\mathbf{p}_c^\text{ref}(t_{LA})$ and the actual landing position $\mathbf{p}_c(t_{LA})$. 

While the proposed jumping framework is aimed primarily at generality and efficiency rather than precision jumping, the data in Fig.~\ref{fig:landing} illustrates that the jumping controller, when using the VBOC, is able to consistently land near its desired final position, with an average error of 11.78$\%$ of the total jumping distance. This is a considerable improvement over the performance over the open loop approach, where the average error is 19.2$\%$ of the total jumping distance. As expected, the error for both approaches increases with jumping distance since friction plays an increasing role as larger tangential forces are needed. In every simulated jump in Fig.~\ref{fig:landing}, the robot landed short of its target. In practice, to account for this  when jumping onto table or over hurdles, we use conservative reference trajectories that allow for a buffer zone between the landing location and the edge of the obstacle.

\subsection{Running Jumps}
\begin{figure}[thb]
    \centering
    \includegraphics[width=\columnwidth]{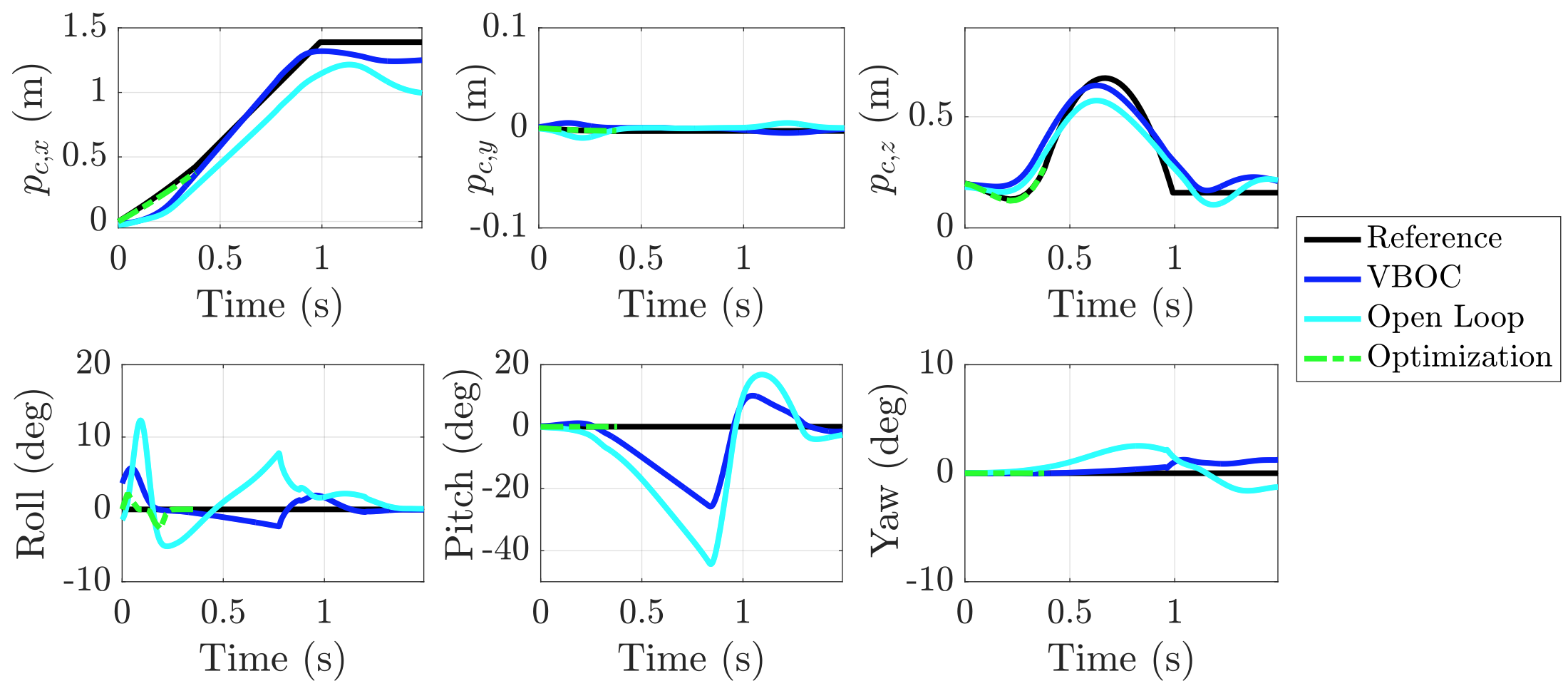}
    \caption{Comparison of reference trajectory tracking performance for the open loop commands versus the VBOC for a trotting jump over a 25~\si{\centi\meter} gap.}
    \label{fig:trot-jump-compare}
\end{figure}

\begin{figure}[thb]
    \centering
    \includegraphics[width=\columnwidth]{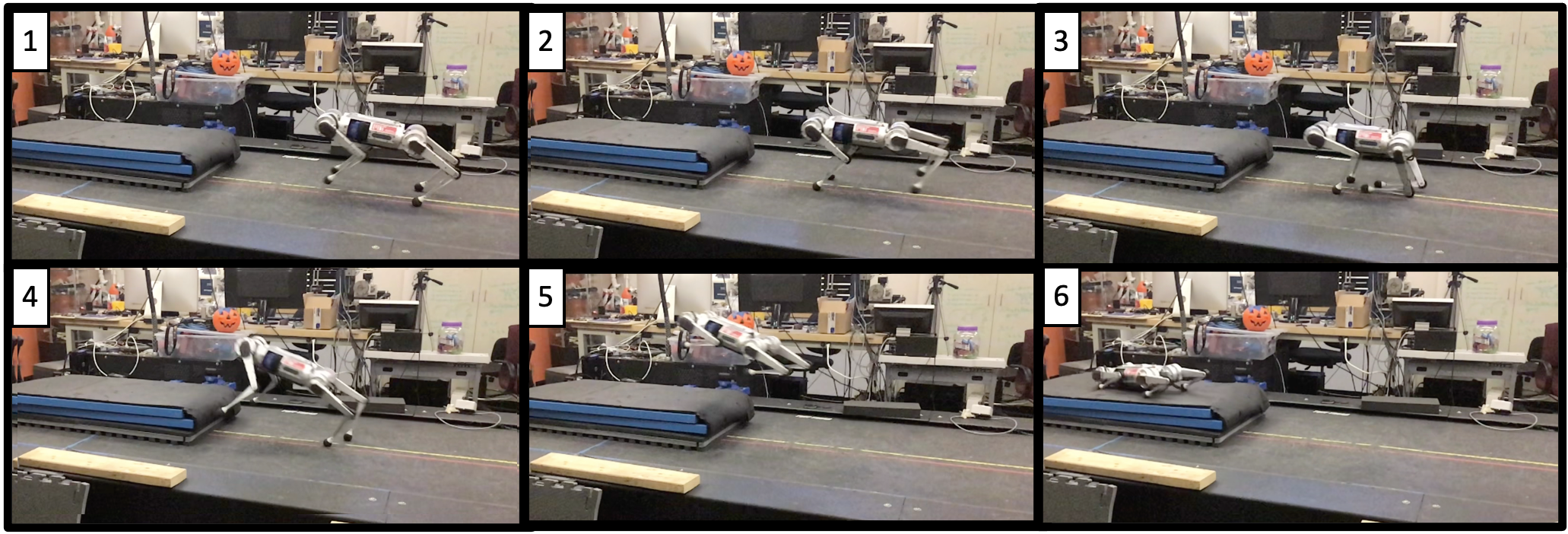}
    \caption{Hardware demonstration of the MIT Mini Cheetah performing a bounding jump onto a 16~\si{\centi\meter} tall platform. The jumping controller takes over starting in the second box.}
    \label{fig:bound-jump-hardware}
\end{figure}

Jumps starting from a dynamic initial state, like those from a running gait, are more complex because the robot will never be exactly in the pose the trajectory optimization expected it to be at the start of takeoff. Simulation results of a trotting jump, shown in Fig.~\ref{fig:trot-jump-compare}, demonstrate that the proposed framework is able not only to generate running jumps given only a contact sequence and a desired landing position, but also to robustly track those jumps in real time. Demonstrations of the trotting and bounding jumps being performed on the hardware are animated in Fig.~\ref{fig:trot-jump} and~\ref{fig:bound-jump-hardware}.

Running jumps have the added challenge of underactuation during takeoff caused by only two feet being in contact at a time. As a result, vertical acceleration of the body is almost always accompanied by some amount of undesired rotation of the body. The tracking controller is able to reason about this underactuation in a robust way that allows the robot to clear obstacles and safely land the jumps. So while the reference trajectory tracking is slightly worse for the running jumps compared to the static jumps, the short solve times of the optimization and the stabilizing capabilities of the tracking controller prove the effectiveness of the framework.

\section{Conclusions}
This paper showed that a framework combining centroidal momentum-based nonlinear optimization with variational-based optimal control can reliably plan and execute rich sets of dynamic aerial motions. Furthermore, we demonstrate that this rich set of novel behaviors emerges from simple cost functions and in the absence of any task specific constraints. The formulation of the trajectory optimization results in short solves times, less than 0.25\si{\second} in worst case, and no observed convergence issues. The compact variational-based optimal controller reasons about underactuation in a way that robustly tracks the vitally important liftoff velocity of the robot. In the future, we plan on extending the framework to robots with an arbitrary number of legs and using the variational-based optimal controller for landing so that the robot can transition back to locomotion immediately upon touching down.

\section*{Acknowledgments}
This work was supported by the Toyota Research Institute, the Centers for ME Research and Education at MIT and SUSTech, and the National Science Foundation Graduate Research Fellowship Program under Grant No. 4000092301.









\bibliographystyle{IEEEtran}
\bibliography{chignoli_bib}

\end{document}